\documentclass[letterpaper, 10 pt, conference]{ieeeconf}
\usepackage{graphicx}
\usepackage{pifont}
\usepackage{booktabs}
\usepackage{makecell}
\usepackage{threeparttable}
\usepackage{algorithmic}
\usepackage{algorithm}
\usepackage{cite}

\IEEEoverridecommandlockouts                              

\title{\LARGE \bf
Efficient Trajectory Generation in 3D Environments with Multi-Level Map Construction
}

\author{Chengkun Tian, Xiaohui Gao and Yonguang Liu
\thanks{Corresponding author:Xiaohui Gao}
\thanks{All authors are with the School of Automation Science and Electrical Engineering,  Beihang University, Beijing, China. $\{$chengkun$\_$tian, gaoxiaohui822,lyg$\}$@buaa.edu.cn.
	    }
	}

\UseRawInputEncoding

\begin{document}

\maketitle
\thispagestyle{empty}
\pagestyle{empty}

\begin{abstract}

We propose a robust and efficient framework to generate global trajectories for ground robots in complex 3D environments. The proposed method takes point cloud as input and efficiently constructs a multi-level map using triangular patches as the basic elements. A kinematic path search is adopted on the patches, where motion primitives on different patches combine to form the global min-time cost initial trajectory. We use a same-level expansion method to locate the nearest obstacle for each trajectory waypoint and construct an objective function with curvature, smoothness and obstacle terms for optimization. We evaluate the method on several complex 3D point cloud maps. Compared to existing methods, our method demonstrates higher robustness to point cloud noise, enabling the generation of high quality trajectory while maintaining high computational efficiency. Our code will be publicly available at https://github.com/ck-tian/MLMC-planner.

\end{abstract}

\section{Introduction}

Ground robots are widely used in exploration and rescue missions within complex 3D environments. Evaluating the multi-level and uneven structures of these environments and generating smooth, collision-free trajectories with kinematic feasibility are crucial for enhancing the autonomy of robots.

A comprehensive map that accurately describes the environment is a prerequisite for effective trajectory generation. Some methods utilize raw point clouds data directly \cite{krusi2017driving}, \cite{liu2015robotic} which provide highly detailed representations of the environments. However, these methods often become computationally intensive due to the complexity in data processing. Voxel maps divide the space into a series of volumetric units known as voxels to represent objects, which are widely used in UAV navigation in 3D environments \cite{zhou2020ego}, \cite{usenko2017real}. However, voxel maps are discrete, making them challenging to accurately represent continuous terrain features such as slopes. Elevation map \cite{fankhauser2018probabilistic} represents environments as 2D height maps, which makes it difficult to represent multi-level structures. Mesh-based \cite{ruetz2019ovpc,putz2021continuous,brandao2020gaitmesh} maps can accurately represent complex structures, but often have high computational complexity and large storage overhead. Trajectory generations using mesh-based maps also have high computational complexity in trajectory smoothing and obstacle detecting.
\begin{figure}[htbp]
	\centering
    \includegraphics[width=\columnwidth]{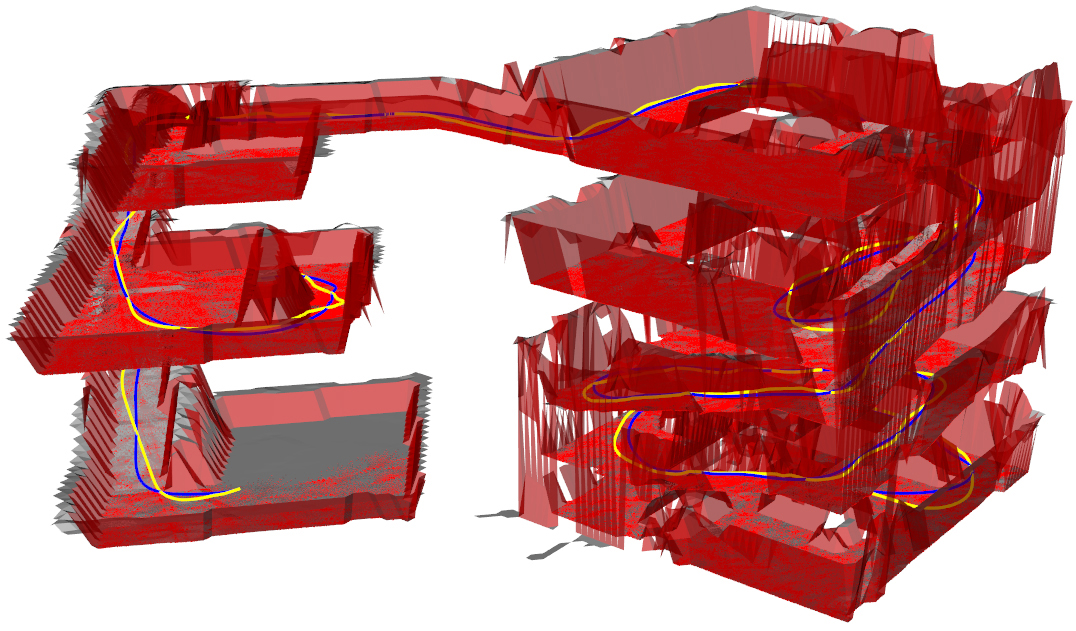}
     \caption{The result of map construction and trajectory generation using the point cloud library (PCL) \cite{rusu20113d} to visualize. The triangular patches divided into traversable  and untraversable are the basic elements of the map. The red dots represent the waypoints of primitives which are generated based on patches. The yellow line represents the inital trajectory consisting of primitives located on traversable patches and the blue line represents the final trajectory after optimization.}
	\label{figurelabel}
\end{figure}

Kinematic search methods \cite{dolgov2010path}, \cite{zhou2019robust} can ensure trajectories smoothness and kinematic feasibility by using motion primitives as graph edges and allow the incorporation of various optimal objectives such as energy and time consumption. Therefore, they are widely adopted in the nonholonomic and multimodal locomotion robots \cite{zhang2022autonomous}. However, generating motion primitives and considering additional constraints are computationally intensive in 3D space.

We propose a robust and highly efficient framework for generating trajectories in complex 3D environments, which includes map construction and trajectory generation. The map construction method uses raw point cloud as input and builds a map using regularly arranged triangular patches, referred to as patches for concision hereafter, as the basic elements. The map construction method is highly computationally efficient and can accurately represents the multi-level and uneven structures. The patches are regularly arranged and share vertices, forming a continuous map. This feature enables efficient indexing, allowing for rapid localization of the robot within the patch and the nearest obstacle at the same level. It significantly enhances computational efficiency of initial trajectory generation and optimization.

We adopt a kinematic path search method to generate motion primitives on the patches. In initial trajectory generation, a method which maintains the smooth of trajectory is proposed for transitioning between different patches. A two-stage optimization is performed to get the final trajectory satisfying ground constraints. Our same-level expansion method guides the trajectory away from obstacles at the current level by excluding the influence from different levels. Our contributions can be summarized as follows:

\begin{figure*}[t]
	\centering
	\includegraphics[width=\textwidth]{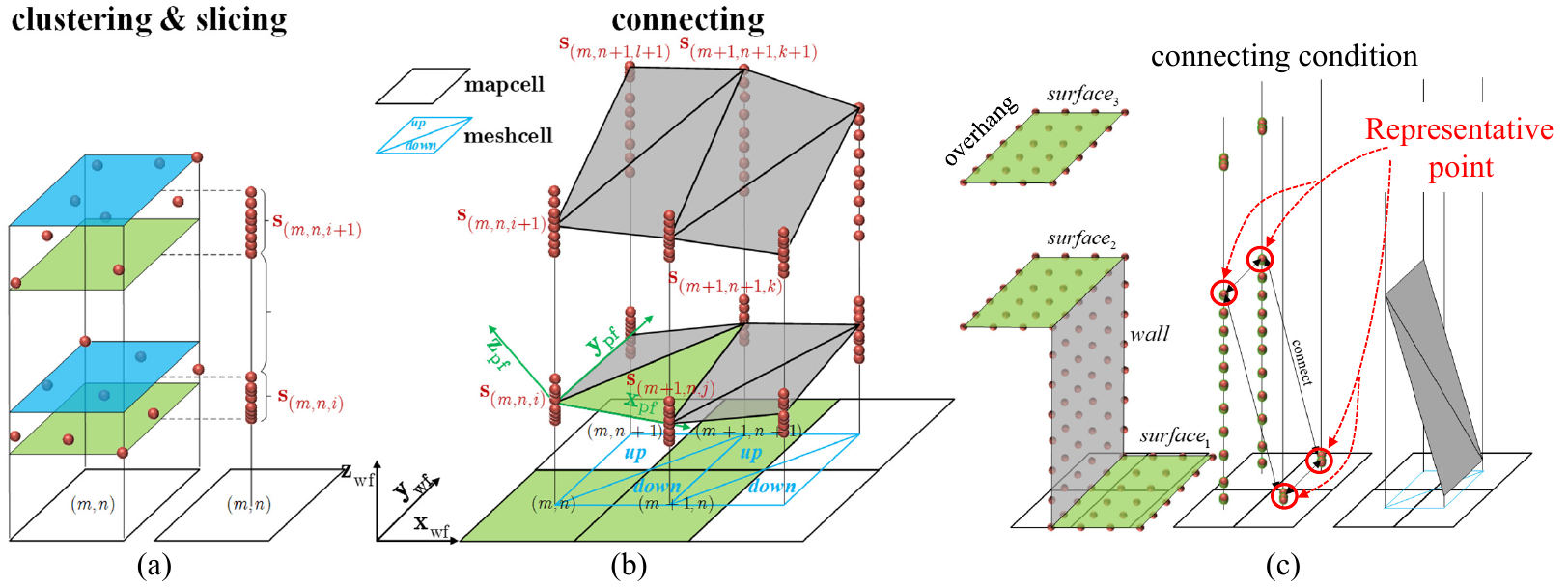}
	\caption{(a) and (b) represent the process of multi-level map construction. ${\rm{(}}{{\bf{x}}_{{\rm{wf}}}},{{\bf{y}}_{{\rm{wf}}}},{{\bf{z}}_{{\rm{wf}}}}{\rm{)}}$ represents the gravity-aligned world frame. ${\rm{(}}{{\bf{x}}_{{\rm{pf}}}},{{\bf{y}}_{{\rm{pf}}}},{{\bf{z}}_{{\rm{pf}}}}{\rm{)}}$ represents the patch frame. Red dots represent the point cloud after clustering. ${{\bf{S}}_{{\rm{(}}m{\rm{,}}n{\rm{,}}i{\rm{)}}}}$ is the $i$-th slice within  ${\bf{mapcel}}{{\bf{l}}_{{\rm{(}}m{\rm{,}}n{\rm{)}}}}$. $\bf{meshcell}$ stores the patches whose projections on the ${{\bf{x}}_{{\rm{wf}}}}$-${{\bf{y}}_{{\rm{wf}}}}$ plane fall within it. In (c), $surface_{\rm{1}}$ is connected to  $surface_{\rm{2}}$ with wall. The $surface_{\rm{3}}$ is an overhang and the height variation from $surface_{\rm{1}}$ to it is not continuous.}
	\label{figurelabe3}
\end{figure*}
1)Efficient Map Construction: We propose an efficient map construction method that uses point cloud as input to construct multi-level map, which can accurately represent the multi-level and uneven structures in complex 3D environment while mitigating the impact of point cloud noise.

2)Kinematic Path Searching: We adopt a kinematic path searching method on triangular patches to generate motion primitives whose transitions between different patches are smooth.

3)Trajectory Optimization: We propose a same-level expansion method to locate the nearest obstacle of the trajectory waypoint at the same level.

\section{Related Works}

\subsection{Map construction}

Triebel et al. \cite{triebel2006multi} propose a Multi-Level Surface Map (MLS Map) representation method. The MLS map stores multiple surfaces within each grid cell and uses intervals to represent vertical structures, thereby providing a comprehensive descriptions of multi-level structures. Nuwan Ganganath et al. \cite{ganganath2015constraint} utilizes grid-based elevation maps to represent terrain.

P{\"u}tz et al.\cite{putz2021continuous} evaluate triangular mesh maps to obtain a layered mesh map and employs the Fast Marching Method (FMM) to compute travel costs through wave-front propagation. Brandao et al. \cite{brandao2020gaitmesh} propose a method for generating navigation meshes in large-scale environments for legged robots. They employ the Ball-Pivoting algorithm \cite{bernardini1999ball} to construct navigation meshes and subsequently assign the most suitable gait controller to each mesh triangle based on the features extracted from the point cloud. Liu \cite{liu2015robotic} proposes a method using raw point clouds as input, leveraging GPU-accelerated tensor voting framework \cite{medioni2000computational} to extract geometric features and using KNN to construct graph nodes.

Wang et al. \cite{wang2023towards} propose a Valid Ground Filter (VGF) to filter raw point clouds, extracting surfaces deemed safe for traversal. Then they construct safety penalty field using the ESDF map and RANSAC fitting. Yang et al. \cite{yang2024efficient} propose a point cloud tomography (PCT) method to generate multi-level 2.5D slices, and further improve computational efficiency through parallel computation. Each slice contains ground and ceiling height layers along with corresponding traversal cost. These methods are computationally efficient, but the noise in point cloud introduces challenges to their performance.

\subsection{Trajectory generation}

Kr{\"u}si et al. \cite{krusi2017driving} generate initial trajectory on raw point cloud using sampling-based method. They assess terrain traversability during motion planning, omitting any kind of explicit map construction. Howard et al. \cite{howard2007optimal} employ numerical linearization and inversion of forward models of propulsion, suspension, and motion to generate trajectory on complex terrain. Some methods use graph search methods, such as Dijkstra \cite{dijkstra2022note} and A*\cite{hart1968formal}, to generate initial paths \cite{brandao2020gaitmesh,wang2023towards,yang2024efficient}. Trajectories can be represented as M-piece 3D polynomials and smoothed in 3D space \cite{wang2023towards} , \cite{yang2024efficient}. Dolgov et al. \cite{dolgov2010path} use non-parametric interpolation to further smooth trajectory, avoiding the oscillation associated with parametric method. 

\section{Multi-Level Map Construction}

This section introduces the method for constructing multi-level map, which consists of three steps: Clustering, Slicing and Connecting.

\subsection{Clustering}

First, the point cloud is clustered according to ${x}$ and ${y}$ coordinates using 2D map grid (Fig.2(a)). Each point ${{\bf{p}}_i} = {({x_i},{y_i},{z_i})}$ belongs to a member of ${\bf{mapcel}}{{\bf{l}}_{(m,n)}}$,
$$
{\rm{s}}{\rm{.t}}{\rm{.}}\left\{ \begin{array}{l}
	m * re{s_{\rm{m}}} - \displaystyle\frac{{re{s_{\rm{m}}}}}{2} \le {x_i} \le m * re{s_{\rm{m}}} + \displaystyle\frac{{re{s_{\rm{m}}}}}{2}\\\\
	n * re{s_{\rm{m}}} - \displaystyle\frac{{re{s_{\rm{m}}}}}{2} \le {y_i}   \le n * re{s_{\rm{m}}} + \displaystyle\frac{{re{s_{\rm{m}}}}}{2}
\end{array} \right.\eqno{(1)}
$$
where ${\bf{mapcel}}{{\bf{l}}_{(m,n)}}$ represents a unit of map grid with index ${\rm{(}}m,n{\rm{)}}$, $re{s_{\rm{m}}}$ represents the size of the map cell.

\subsection{Slicing}

The points in the map cell are sliced according to their $z$ coordinates (Fig.2(a)). The slicing method is similar to \cite{triebel2006multi}. Given the $z$ values presorted in ascending order, each $z$ value is compared with the next one. For the current value ${z_i}$ and the next value ${z_{i + {\rm{1}}}}$ , if ${z_{i + {\rm{1}}}} - {z_i} > th{r_{{\rm{slice}}}}$ ( $th{r_{{\rm{slice}}}}$ is the minimum slice gap, which should be greater than the height of the robot), then ${z_i}$ represents the end of the current slice and ${z_{i + {\rm{1}}}}$ represents the start of the next slice. Slices containing fewer points than $least\_num$ will be discarded, which filters out noise in the point cloud. After obtaining a valid slice, we compute the average (${z_{{\rm{avg}}}}$), maximum (${z_{{\rm{max}}}}$), and minimum (${z_{{\rm{min}}}}$) values for all the $z$ values in the slice.

\subsection{Connecting}

Three adjacent slices within different map cells are connected to form the basic element of the map: patch. As shown in Fig.2(b), given ${\bf{mapcel}}{{\bf{l}}_{\left( {m{\rm{,}}n} \right)}}$, the connections are established in two search pairs: ${\bf{mapcell}}{{\bf{s}}_{\left( {m{\rm{,}}n} \right),\left( {m + 1{\rm{,}}n} \right),\left( {m + 1{\rm{,}}n + 1} \right)}}$ and ${\bf{mapcell}}{{\bf{s}}_{\left( {m{\rm{,}}n} \right),\left( {m + 1{\rm{,}}n + 1} \right),\left( {m{\rm{,}}n + 1} \right)}}$. The condition of connecting two slices is determined by their distance in the vertical direction (gravity-aligned). Specifically, given two slices, ${{\bf{S}}_{\rm{1}}}$ and ${{\bf{S}}_{\rm{2}}}$, the condition for them to be connected is:
$$
\begin{array}{l}
	({z_{{\rm{1min}}}} - {z_{{\rm{2max}}}} \le \lambda  * re{s_{{\rm{pc}}}})\;{\rm{and}}\\[1ex]
	({z_{{\rm{2min}}}} - {z_{{\rm{1max}}}} \le \lambda  * re{s_{{\rm{pc}}}})
\end{array}\eqno{(2)}
$$
where $z_{{\rm{1min}}}$ and $z_{{\rm{1max}}}$ are the minimum and maximum values of the ${{\bf{S}}_{\rm{1}}}$ respectively, $re{s_{{\rm{pc}}}}$ is the resolution of the point cloud, $\lambda$ accounts for the irregular distribution of the point cloud and should exceed 1.0. Two slices satisfying (2) indicates that their vertical distance does not exceed $\lambda  * re{s_{{\rm{pc}}}}$.

As shown in Fig.2(c), this condition is based on the observation that height variations in 3D space are divided into continuous (with connection through physical entities such as walls or slopes) and discontinuous (without physical entities, such as overhangs). When two adjacent slices correspond to regions with continuous height variations, they capture the point clouds of the physical entities. This feature is reflected in the span of these slices, where their vertical distance does not exceed the resolution of the point cloud.

A patch is defined by three points in 3D space. Therefore, each slice needs to be represented by a single point, referred to as representative point. The $x$ and $y$ values of the representative point are determined by the index of map cell. The following discusses the $z$ value of it.

As shown in Fig. 3, using the average values to represent slices corresponding to regions with gentle slope variation is reasonable. Using the maximum values in these slices may introduce biases in the vertical direction. However, for slices corresponding to regions with steep slope variation, using the mean values may result in losses of height information. The slices corresponding to the regions with steep slope variation have larger spans. The method to determine the $z$ value is:
$$
z = \left\{ \begin{array}{l}
	{z_{{\rm{max}}}} \quad  \left( {{z_{{\rm{max}}}} - {z_{{\rm{min}}}}} \right) > th{r_{{\rm{rep}}}}\\[1ex]
	{z_{{\rm{avg}}}} \quad\;  \left( {{z_{{\rm{max}}}} - {z_{{\rm{min}}}}} \right) \le th{r_{{\rm{rep}}}}
\end{array} \right.\eqno{(3)}
$$
where $th{r_{{\rm{rep}}}}$ is a value for classification.

As shown in Fig. 2(b), connecting pairs \{${{\bf{S}}_{(m,n,i)}}$ , ${{\bf{S}}_{{\rm{(}}m + 1{\rm{,}}n{\rm{,}}j{\rm{)}}}}$\} and \{${{\bf{S}}_{{\rm{(}}m + 1{\rm{,}}n{\rm{,}}j{\rm{)}}}}$ , ${{\bf{S}}_{{\rm{(}}m + 1{\rm{,}}n + 1{\rm{,}}k{\rm{)}}}}$\} which belong to the search pair ${\bf{mapcell}}{{\bf{s}}_{\left( {m{\rm{,}}n} \right),\left( {m + 1{\rm{,}}n} \right),\left( {m + 1{\rm{,}}n + 1} \right)}}$ satisfy (2). These three slices use the average values as the ${z}$ values of their representative points to form the green patch.
Connecting pairs \{${{\bf{S}}_{{\rm{(}}m{\rm{,}}n{\rm{,}}i+1{\rm{)}}}}$ , ${{\bf{S}}_{{\rm{(}}m{\rm{,}}n + 1{\rm{,}}l + 1{\rm{)}}}}$\} and \{(${{\bf{S}}_{{\rm{(}}m{\rm{,}}n + 1{\rm{,}}l + 1{\rm{)}}}}$ ,  ${{\bf{S}}_{{\rm{(}}m + 1{\rm{,}}n + 1{\rm{,}}k + 1{\rm{)}}}}$)\} which belong to the search pair ${\bf{mapcell}}{{\bf{s}}_{\left( {m{\rm{,}}n} \right),\left( {m + 1{\rm{,}}n + 1} \right),\left( {m{\rm{,}}n + 1} \right)}}$ also satisfy (2), with the ${{\bf{S}}_{{\rm{(}}m{\rm{,}}n + 1{\rm{,}}l + 1{\rm{)}}}}$ and ${{\bf{S}}_{{\rm{(}}m + 1{\rm{,}}n + 1{\rm{,}}k + 1{\rm{)}}}}$ using the maximum values. The patch they form, along with the previously described green patch, both belong to ${\bf{meshcel}}{{\bf{l}}_{{\rm{(}}m{\rm{,}}n{\rm{)}}}}$. The ${\bf{meshcel}}{{\bf{l}}_{{\rm{(}}m{\rm{,}}n{\rm{)}}}}$ is a unit of mesh grid with index ${\rm{(}}m,n{\rm{)}}$ and has the same shape as the $\bf{mapcell}_{(m,n)}$, with a positional offset of $0.5re{s_{\rm{m}}}$. The two patches's projections on the ${{\bf{x}}_{{\rm{wf}}}}$-${{\bf{y}}_{{\rm{wf}}}}$ plane occupy different parts of the ${\bf{meshcel}}{{\bf{l}}_{{\rm{(}}m{\rm{,}}n{\rm{)}}}}$ which is further divided into ${\bf{meshcell}}_{{\rm{(}}m{\rm{,}}n{\rm{)}}}^{{\rm{up}}}$ and ${\bf{meshcell}}_{{\rm{(}}m{\rm{,}}n{\rm{)}}}^{{\rm{down}}}$.
  
\begin{figure}[t]
	\centering
	\includegraphics[width=\columnwidth]{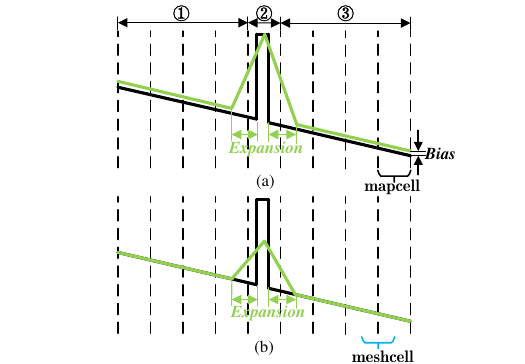}
	\caption{The black lines represent the raw point cloud data. The green lines represent the results of map construction. The regions between two adjacent dashed lines represent map cells. In (a), each slice uses ${z_{{\rm{max}}}}$ as the $z$ value of the representative point, which results in a vertical direction bias in regions with gentle slope variation, such as regions \ding{172} and \ding{174}. In (b), each slice uses ${z_{{\rm{avg}}}}$. While there is no vertical direction bias in regions with gentle slope variation, the height information in regions with steep slope variation, such as area \ding{173}, is lost.}
	\label{figurelabe4}
\end{figure}

Each patch integrates all point cloud information within the corresponding height ranges of three map cells (as the patch and map cells marked green shown in Fig. 2(b)), which mitigates the impact of point cloud noise. The patches have regular projection shapes and the mesh cells are further divided, enabling rapid indexing. Additionally, as shown in Fig. 3 and Fig. 7, the proposed map construction method inflates convex obstacles by ${\rm{(0}}{\rm{.5}}re{s_{{\rm{m}}}} \sim {\rm{1}}{\rm{.5}}re{s_{{\rm{m}}}})$, with the exact size determined by the relative positions of the map cells to the obstacles. Considering the safe margin, $r<0.5re{s_{{\rm{m}}}}$ should be satisfied, where $r$ is the minimum radius of the sphere enclosing robot. Under this condition, the initial trajectory generated satisfies the collision-free constraint, however, it restricts the patch size, thus reducing the map representation accuracy. In practical experiments, the safe margin can be disregarded during map construction, and the collision-free constraint can instead be addressed in the trajectory optimization phase.

During the connecting step, the patches located in the same part of the same mesh cell cannot share vertices. If multiple patches share vertices, only the uppermost one is retained. This method avoids cross connections in regions with complex structures, ensuring that the distance between the patches at different levels is not less than $th{r_{{\rm{slice}}}}$.

After obtaining a patch, a patch frame ${\rm{(}}{{\bf{x}}_{{\rm{pf}}}}{\rm{,}}{{\bf{y}}_{{\rm{pf}}}}{\rm{,}}{{\bf{z}}_{{\rm{pf}}}}{\rm{)}}$ is constructed as shown in Fig. 2.
\begin{figure}[t]
	\centering
	\includegraphics[width=\columnwidth]{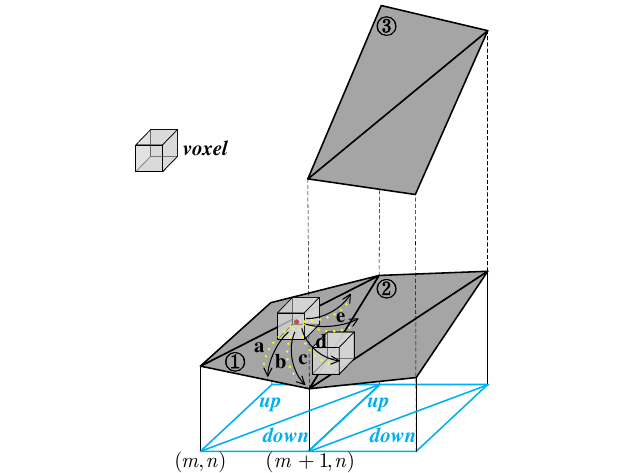}
	\caption{The red dot represents the graph node currently being expanded and the yellow dots lines represent the motion primitives. The motion primitive ${\bf{c}}$ transitions from  patch \ding{172} to \ding{173} and expands to a new graph node. The voxels discretize the 3D space and each voxel contains only one graph node by pruning.  }
	\label{figurelabe5}
\end{figure}
$$
\left\{ \begin{array}{l}
	{{\bf{z}}_{{\rm{pf}}}} = {\bf{n}} = {\rm{(}}{n_1}{\rm{,}}{n_2}{\rm{,}}{n_3}{\rm{)}} =\displaystyle \frac{{\left( {{\bf{b}} - {\bf{a}}} \right) \times \left( {{\bf{c}} - {\bf{a}}} \right)}}{{\left\| {\left( {{\bf{b}} - {\bf{a}}} \right) \times \left( {{\bf{c}} - {\bf{a}}} \right)} \right\|}}\\[1ex]
	{{\bf{x}}_{{\rm{pf}}}} = \displaystyle\frac{{\left( {{n_3}{\rm{,}}0{\rm{,}} - {n_1}} \right)}}{{\left\| {\left( {{n_3}{\rm{,}}0{\rm{,}} - {n_1}} \right)} \right\|}}\\[1ex]
	{{\bf{y}}_{{\rm{pf}}}} = \displaystyle\frac{{{{\bf{z}}_{{\rm{pf}}}} \times {{\bf{x}}_{{\rm{pf}}}}}}{{\left\| {{{\bf{z}}_{{\rm{pf}}}} \times {{\bf{x}}_{{\rm{pf}}}}} \right\|}}
\end{array} \right.\eqno{(4)}
$$
where ${\bf{a}}$, ${\bf{b}}$, ${\bf{c}}$ are the three vertices of the patch. The ${\bf{a}}$ is the origin of the patch frame, the patch frame satisfies ${n_3} > 0$ and ${{\bf{x}}_{{\rm{pf}}}} \bot {{\bf{y}}_{{\rm{wf}}}}$.

Vector ${\bf{n}}$ represents the slope of the patch and classifies the patch into traversable and untraversable. The constructed map only retains traversable patches.

The transformation matrix of the patch frame relative to the world frame is given by:
$$
{\bf{T}} = \left[ {\begin{array}{*{20}{c}}
		{{{\bf{x}}_{{\rm{pf}}}}^T}&{{{\bf{y}}_{{\rm{pf}}}}^T}&{{{\bf{z}}_{{\rm{pf}}}}^{\mathop{\rm T}\nolimits} }&{{{\bf{a}}^T}}\\[1ex]
		0&0&0&1
\end{array}} \right]\eqno{(5)}
$$
We evaluate the accuracy of the map construction result in Section V, Part B.

\section{Trajectory Generation}

This section introduces the method of trajectory generation. Initially, we discuss the local trajectory generation on a single patch and transitions between different patches. During trajectory optimization, we discuss the same-level expansion method, which identifies the nearest obstacle at the same level.

\begin{algorithm}[!h]
	\caption{Algorithm of Expanding Node}
	\label{alg:2}
	\renewcommand{\algorithmicrequire}{\textbf{Input:}}
    \renewcommand{\algorithmicensure}{\textbf{Output:}}
	\begin{algorithmic}[1]
		\REQUIRE $n_p$  
		\ENSURE $nodes$    
		\STATE $w_0 \leftarrow \textbf{GetState($n_p$)}$
		\FOR{${{\bf{u}}_i}  \in  \mathcal{U}$}
		\STATE $k\leftarrow 1$
		\STATE $primitive.\textbf{clear}()$
		\FOR{$k\ \textbf{to} \ \it{nu{m_{iter}}} $}
		\STATE ${w_k} \leftarrow \textbf{StateTransition}({{\bf{u}}_i},{w_{k-1}})$
		\IF{$\textbf{ExitPatch}({w_k})$}
		\STATE $success,patch_{c} \leftarrow \textbf{SearchPatch}({w_k})$
		\IF{$success$}
		\STATE ${w_k} \leftarrow \textbf{AdjustState}({w_k,patch_c})$
		\ELSE
		\STATE $ \textbf{break} $
		\ENDIF
		\ENDIF
		\STATE $ primitive.\textbf{add}({w_k})$
		\ENDFOR
		\IF{$k==\it{nu{m_{iter}}}$}
		\STATE $ n_c \leftarrow \textbf{CreatNode}({\bf{u}_i},{primitive}) $
	\STATE $nodes.\textbf{add}(n_c)$
	\ENDIF
	\ENDFOR	
	\RETURN $nodes$
\end{algorithmic}
\end{algorithm}

\subsection{Initial trajectory generating}

Our method to generate the loacal trajectory originates from the hybride-state A* \cite{dolgov2010path} and the fast-planner \cite{zhou2019robust}, which use motion primitives as graph edges and perform a graph search loop similar to the A* algorithm \cite{hart1968formal}. 

As shown in Fig. 4 and Alg. 1, parent node $n_p$ is expanded through generating motion primitives, resulting in child nodes $nodes$. The primitives are sequences of waypoints \{$ w_0,w_1,\cdots,w_{num_{iter}}$\} which connect the graph nodes. Given the control input and the current waypoint, the next waypoint is obtained through \textbf{StateTransition()}. Each waypoint maintains the robot's state and the patch located on. \textbf{ExitPatch()} checks whether the newly generated waypoint exits the previous patch, then \textbf{SearchPatch()} search the patch currently located on. The failure of \textbf{SearchPatch()} implies the waypoint encounters a collision with untraversable region and the corresponding control input should be discarded.

\begin{figure}[t]
	\centering
	\includegraphics[width=\columnwidth]{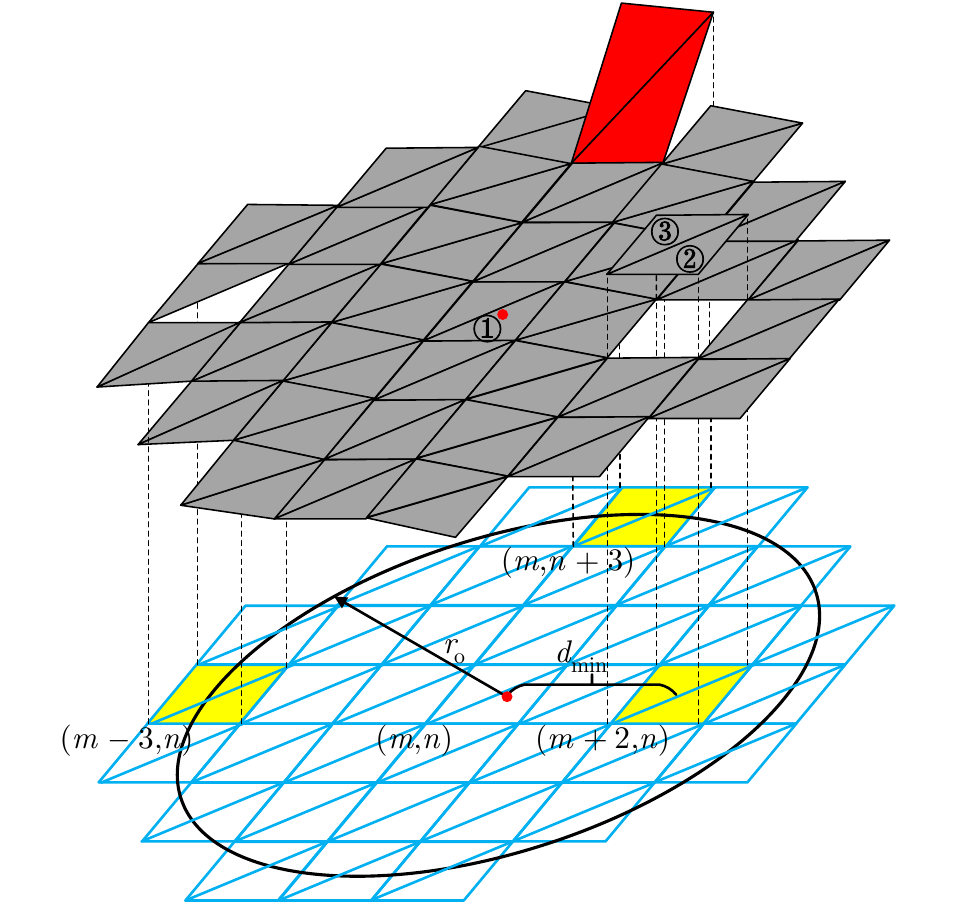}
	\caption{ The red dot represents a trajectory waypoint located on patch \ding{172}. Except for patch \ding{173} and \ding{174}, all other patches are at the same level as the patch \ding{172}. Mesh cells marked in yellow are obstacles identified through the same-level expansion. Both parts of ${\bf{meshcel}}{{\bf{l}}_{{\rm{(}}m{\rm{,}}n + 3{\rm{)}}}}$ contain a patch at the same level while untraversable. Both parts of  ${\bf{meshcel}}{{\bf{l}}_{{\rm{(}}m + 2{\rm{,}}n{\rm{)}}}}$ don't contain patches at the same level. Only one part of ${\bf{meshcel}}{{\bf{l}}_{{\rm{(}}m - 3{\rm{,}}n{\rm{)}}}}$ contains traversable patch at the same level. The nearest obstacle to the trajectory waypoint is the ${\bf{meshcel}}{{\bf{l}}_{{\rm{(}}m + 2{\rm{,}}n{\rm{)}}}}$.}
	\label{figurelabe6}
\end{figure}

Consider the kinematic model of a differential drive robot navigating on a patch:
$$
\left\{ \begin{array}{l}
	{{\dot x}_{{\rm{local}}}} = \cos \theta  * \displaystyle\frac{{{v_{{\rm{left}}}} + {v_{{\rm{right}}}}}}{{\rm{2}}}\\[1ex]
	{{\dot y}_{{\rm{local}}}} = \sin \theta  * \displaystyle\frac{{{v_{{\rm{left}}}} + {v_{{\rm{right}}}}}}{{\rm{2}}}\\[1ex]
	{{\dot z}_{{\rm{local}}}} = {\rm{0}}\\[1ex]
	{{\dot v}_{{\rm{left}}}}  = {a_{{\rm{left}}}}\\[1ex]
	{{\dot v}_{{\rm{right}}}} = {a_{{\rm{right}}}}\\[1ex]
	\dot \theta  = \displaystyle\frac{{{v_{{\rm{left}}}} - {v_{{\rm{right}}}}}}{l}
\end{array} \right.\eqno{(6)}
$$
where ${\rm{(}}{x_{{\rm{local}}}}{\rm{,}}{y_{{\rm{local}}}}{\rm{,}}{z_{{\rm{local}}}}{\rm{)}}$ is the position of the robot in the patch frame, ${v_{{\rm{left}}}}$ and ${v_{{\rm{right}}}}$ are the linear velocities of the left and right wheels respectively, $\theta$ is the robot's orientation angle in the patch frame, $l$ is the track width and ${a_{{\rm{left}}}}$ and  ${a_{{\rm{right}}}}$ are the linear accelerations of the left and right wheels respectively.
$$
\left( \begin{array}{l}
	{{\dot x}_{{\rm{global}}}}\\[1ex]
	{{\dot y}_{{\rm{global}}}}\\[1ex]
	{{\dot z}_{{\rm{global}}}}\\[1ex]
\end{array} \right) = {{\bf{T}}_{\left[ {1:3,1:3} \right]}}\left( \begin{array}{l}
	{{\dot x}_{{\rm{local}}}}\\[1ex]
	{{\dot y}_{{\rm{local}}}}\\[1ex]
	{{\dot z}_{{\rm{local}}}}
\end{array} \right)\eqno{(7)}
$$
where $({x_{{\rm{global}}}},{y_{{\rm{global}}}},{z_{{\rm{global}}}})$ is the position of the robot in the world frame and ${{\bf{T}}_{\left[ {1:n,1:n} \right]}}$ is the top-left $n \times n$ submatrix of ${\bf{T}}$ (using 1-based indexing).

 ${\bf{x}}={{\rm{(}}{x_{{\rm{global}}}}{\rm{,}}{y_{{\rm{global}}}}{\rm{,}}{z_{{\rm{global}}}}{\rm{,}}{v_{{\rm{left}}}}{\rm{,}}{v_{{\rm{right}}}}{\rm{,}}\theta {\rm{)}}}$ is selected as the state vector and ${\bf{u}}= {{\rm{(}}{a_{{\rm{left}}}}{\rm{,}}{a_{{\rm{right}}}}{\rm{)}}}$ as the control input. Then $w=\{{\bf{x}},patch\}$ and ${{\bf{u}}}  \in  \mathcal{U}$ which is discretized within the range $[ - {a_{{\rm{max}}}},{a_{\max }}]$. The equations used  in \textbf{StateTransition()} can be derived from (6),(7):
 $$
\left\{ \begin{array}{l}
	\Delta {v_{{\rm{left}}}} = {a_{{\rm{lef}}t}}\Delta t\\
	\Delta {v_{{\rm{right}}}} = {a_{{\rm{right}}}}\Delta t\\
	\Delta \theta  = \displaystyle\frac{{{v_{{\rm{left}},n - 1}} - {v_{{\rm{right}},n - 1}}}}{l}\Delta t + \displaystyle\frac{{{a_{{\rm{lef}}t}} - {a_{{\rm{right}}}}}}{{2l}}{(\Delta t)^2}\\
	\Delta {x_{{\rm{global}}}} = \displaystyle\frac{{{v_{n - 1}}\cos ({\theta _{n - 1}}) + {v_n}\cos ({\theta _n})}}{2}\Delta t{T_{11}}\ \ \ \ \ \ (8)\\
	+ \displaystyle\frac{{{v_{n - 1}}\sin ({\theta _{n - 1}}) + {v_n}\sin ({\theta _n})}}{2}\Delta t{T_{12}}\\
	\Delta {y_{{\rm{global}}}} = \displaystyle\frac{{{v_{n - 1}}\cos ({\theta _{n - 1}}) + {v_n}\cos ({\theta _n})}}{2}\Delta t{T_{21}}\\
	+ \displaystyle\frac{{{v_{n - 1}}\sin ({\theta _{n - 1}}) + {v_n}\sin ({\theta _n})}}{2}\Delta t{T_{22}}
\end{array} \right.
 $$ 
 where $\Delta\theta$ is ${\theta _n}-{\theta _{n - 1}}$, and so on, ${v}$ is the central velocity given by $({v_{{\rm{left}}}} + {v_{{\rm{right}}}})/2$ and ${T_{ij}}$ is an element of ${\bf{T}}$ at ${i}$-th row and ${j}$-th column. The components of ${v}$ in the patch frame are approximated as linearly varying over $\Delta t$.  

The condition for waypoint transition between two patches is that they share vertices. Since the arrangement of the patches is regular, given the ${\rm{(}}{x_{{\rm{global}}}}{\rm{,}}{y_{{\rm{global}}}}{\rm{)}}$ of current waypoint and the patch which previous waypoint located on, the current patch located on can be quickly determined (performed in \textbf{ExitPatch()} and \textbf{SearchPatch()}). As the motion primitive ${\bf{c}}$ shown in Fig. 4, the waypoints located on patch \ding{172} within ${\bf{meshcell}}_{{\rm{(}}m{\rm{,}}n{\rm{)}}}^{{\rm{down}}}$ satisfy:
$$
\left\{ \begin{array}{l}
	m * re{s_{\rm{m}}} \le {x_{{\rm{global}}}} \le \left( {m + 1} \right) * re{s_{\rm{m}}}\\[1ex]
	n * re{s_{\rm{m}}} \le {y_{{\rm{global}}}} \le \left( {n + 1} \right) * re{s_{\rm{m}}}\\[1ex]
	{y_{{\rm{global}}}} - n * re{s_{\rm{m}}} \le {x_{{\rm{global}}}} - m * re{s_{\rm{m}}}
\end{array} \right.\eqno{(9)}
$$
while the last waypoint in the motion primitive ${\bf{c}}$ satisfies:
$$
\left\{ \begin{array}{l}
	(m + {\rm{1}}) * re{s_{\rm{m}}} \le {x_{{\rm{global}}}} \le \left( {m + {\rm{2}}} \right) * re{s_{\rm{m}}}\\[1ex]
	n * re{s_{\rm{m}}} \le {y_{{\rm{global}}}} \le \left( {n + {\rm{1}}} \right) * re{s_{\rm{m}}}\\[1ex]
	{y_{{\rm{global}}}} - n  * re{s_{\rm{m}}} > {x_{{\rm{global}}}} - (m + {\rm{1}}) * re{s_{\rm{m}}}
\end{array} \right.\eqno{(10)}
$$
This means the last waypoint exits the patch \ding{172} and is located in the ${\bf{meshcell}}_{{\rm{(}}m + 1{\rm{,}}n{\rm{)}}}^{{\rm{up}}}$. The patch \ding{173} within the ${\bf{meshcell}}_{(m + 1,n)}^{{\mathop{\rm up}\nolimits} }$ shares  vertices with the patch \ding{172}. Therefore, the last waypoint is located on the patch \ding{173}.

When transitioning to a different patch, the robot's orientation and ${z_{{\rm{global}}}}$ need to be adjusted (performed in \textbf{AdjustState()}), as they are constrained by the patch located on. The adjusted orientation is obtained by:
$$
k\left( \begin{array}{l}
	\cos {\theta _{{\rm{cm}}}}\\
	\sin {\theta _{{\rm{cm}}}}
\end{array} \right) = {\bf{T}}_{{\rm{cm}}[1:2{\rm{,}}1:2]}^T{{\bf{T}}_{{\rm{pm}}}}_{[1:2{\rm{,}}1:2]}\left( \begin{array}{l}
	\cos {\theta _{{\rm{pm}}}}\\
	\sin {\theta _{{\rm{pm}}}}
\end{array}\right)(11)
$$
where ${\theta _{{\rm{pm}}}}$ is the orientation before adjustment, ${\theta _{{\rm{cm}}}}$ is the adjusted orientation, ${{\bf{T}}_{{\rm{pm}}}}$ and ${{\bf{T}}_{{\rm{cm}}}}$ are the transformation matrices of the previous and the current patches respectively and $k$ is a scaling factor.

This adjustment method maintains the consistency of the orientations' projections on the ${{\bf{x}}_{{\rm{wf}}}}$-${{\bf{y}}_{{\rm{wf}}}}$ plane. Although not fully conforming to the contact model between robot and supporting ground, it ensures the smoothness of the trajectory while maintaining high computational efficiency.

Then the adjusted ${z_{{\rm{global}}}}$ is obtained by:
$$
{\bf{T}}_{{\rm{cm}}[3{\rm{,:}}]}^{ - 1}{[\begin{array}{*{20}{c}}
		{{x_{{\rm{global}}}}}&{{y_{{\rm{global}}}}}&{{z_{{\rm{global}}}}}&1
	\end{array}]^T} = 0\eqno{(12)}
$$
where ${\bf{T}}_{{\rm{cm}}[3{\rm{,}}:]}^{ - 1}$ is the third row of ${\bf{T}}_{{\rm{cm}}}^{ - 1}$.

We don't explicitly assign travel cost to each patch. In the A* search loop, the edge cost is time comsumed $t_p$, and the heuristic is $ {\parallel} {\bf{Pos}_{\rm{goal}}} - {\bf{Pos}_{\rm{cur}}} {\parallel} / {v_{\rm{max}}}$ which is calculated as the Euclidean distance between the current and goal position divided by the maximum velocity and represents the lower bound of the trajectory cost. The min-time search strategy guides the trajectory along  patches with lower slopes. 
\subsection{Trajectory Optimization}

Same to the hybrid-state A* algorithm [11], we construct an objective function (13) for the initial trajectory based on collision, curvature, and smoothness, then optimize it using conjugate-gradient (CG) descent.
$$
\begin{array}{l}
	{\omega _{\rm{o}}}\sum\limits_{i = 1}^N {{\sigma _{\rm{o}}}(|{{\bf{x}}_i} - {{\bf{o}}_i}} | - {r_{\rm{o}}}) + {\omega _{\rm{c}}}\sum\limits_{i = 1}^{N - 1} {{\sigma _{\rm{c}}}(\frac{{\Delta {\phi _i}}}{{|\Delta {x_i}|}}}  - {c_{{\rm{max}}}})\\
	+ {\omega _{\rm{s}}}\sum\limits_{i = 1}^{N - 1} {{{(\Delta {{\bf{x}}_{i + 1}} - \Delta {{\bf{x}}_i})}^2}} 
\end{array} (13)
$$
where ${{\bf{x}}_i}={\rm{(}}{x_i}{\rm{,}}{y_i}{\rm{)}}$ is the ${\rm{(}}{x_{{\rm{global}}}}{\rm{,}}{y_{{\rm{global}}}}{\rm{)}}$ of the $i$-st trajectory waypoint, ${{\bf{o}}_i}$ is the 2D location of the nearest obstacle to the $i$-st trajectory waypoint. $\Delta {{\bf{x}}_i} = {{\bf{x}}_i} - {{\bf{x}}_{i - 1}}$.
$\Delta \phi  = |{\tan ^{ - 1}}\frac{{\Delta {y_{i + 1}}}}{{\Delta {x_{i + 1}}}} - {\tan ^{ - 1}}\frac{{\Delta {y_i}}}{{\Delta {x_i}}}|$.

Equation (13) is only based on the $x$ and $y$ values of the trajectory waypoints, as the $z$ values are constrained by the patches located on. Given the ${\rm{(}}{x_{{\rm{global}}}}{\rm{,}}{y_{{\rm{global}}}}{\rm{)}}$ and the patch located on, the ${z_{{\rm{global}}}}$ is uniquely determine. The same methods of the waypoint transition between patches and the ${z_{{\rm{global}}}}$ adjustment as in Section IV, Part A are adopted during iterations.

The nearest obstacle found using the ESDF map or the Voronoi diagram may not be at the same level as the waypoint. 
As shown in Fig. 5, we employ a same-level expansion method to identify the nearest obstacles: Using the current waypoint as center and ${r_{\rm{o}}}$ as radius, this method searches for same-level patches within this area. Same-level means that two patches can be connected through already expanded patches by sharing vertices. A mesh cell is not considered an obstacle only if both parts contain traversable patches at the same level.

After completing the above optimization iterations, linear interpolations between the trajectory waypoints are adopt to obtain a sequence of waypoints with smaller intervals. Then an objective function with 3D curvature and smoothness terms (based on the $x$, $y$ and $z$ values) similar to (13) is constructed for this sequence, resulting in the final trajectory. To avoid deviation from the ground constraint, fixng the original waypoints or fewer iterations are adopted.

\section{Experiment}

\subsection{Implementation Details}
\begin{table}[h]
	\centering
	\begin{threeparttable}
		
		\caption{Scenario Parameters}
		\label{table1}
		\begin{tabular}{c|ccc}
			\toprule
			\textit{Scenario} & \textbf{$l \times w \times h({{\mathop{\rm m}\nolimits} ^{\rm{3}}})$} & \textbf{$num$} & \textbf{$re{s_{{\rm{pc}}}}({\mathop{\rm m}\nolimits} )$}  \\ \midrule
			\textit{Spiral} \cite{wang2023towards} & 78 $\times$ 36 $\times$ 22 & 231885 & 0.2 \\
			\textit{Uneven terrain} & 111 $\times$ 111 $\times$ 57 & 624001 & 0.2 \\
			\textit{Building} & 31 $\times$ 33 $\times$ 18 & 828259 & 0.1 \\
			\bottomrule
		\end{tabular}
		\begin{tablenotes} 
			\item$l \times w \times h$ is the dimension of scenario, $num$ is the number of points in point cloud and $re{s_{{\rm{pc}}}}$ is the resolution of point cloud.
		\end{tablenotes} 
	\end{threeparttable}
\end{table}
We evaluate the map construction and trajectory generation methods in three scenarios with complex 3D structures:

1)\textit{Spiral}\cite{wang2023towards}: The point cloud map in \cite{wang2023towards} consists of two spiral roadways and a bridge without guardrails (Fig. 6 (a1)).

2)\textit{Uneven terrain}: The point cloud map consists of two layers of uneven terrain connected by a spiral roadway (Fig. 6 (b1).

3)\textit{Building}: The point cloud map of a complex indoor scenario with five layers (Fig. 7 (a1)). Unlike the other two scenarios, we construct this scenario in Gazebo \cite{koenig2004design} and generate its point cloud by fusing multiple frames of simulated RGB-D camera scans from different poses. This method enables the experiment to simulate the irregular density of large scale point cloud and the point cloud drift caused by pose estimation inaccuracy (Fig. 7 (b1, b2)). 
The $res_{\rm{m}}$ should be greater than the $res_{\rm{pc}}$ to ensure the formation of the slice, which is a prerequisite for the connecting, and we choose $res_{\rm{m}}$=$3res_{\rm{pc}}$ in this experiment.

\begin{table}[t]
	\centering
	\begin{threeparttable}
		
		\caption{Map Construction Accuracy}
		\label{table2}
		\begin{tabular}{c|cccc}
			\toprule
			\textit{Scenario} & \textbf{$re{s_{{\rm{m}}}}({\mathop{\rm m}\nolimits} )$} & \textbf{$th{r_{{\rm{slope}}}}({\rm{^\circ }})$} & \textbf{$nu{m_{\rm{t}}}$} & \textbf{${E_{{\rm{avg}}}}({\mathop{\rm m}\nolimits} )$} \\ \midrule
			\textit{Spiral} \cite{wang2023towards} &0.6 & 40 & 26832 & 0.093 \\
			\textit{Uneven terrain} &0.6 & 40 & 107258 & 0.022 \\
			\textit{Building} &0.3 & 40 & 22354 & 0.034 \\

			\bottomrule
		\end{tabular}
		\begin{tablenotes} 
			\item Patches with slopes below $th{r_{{\rm{slope}}}}$ are considered traversable, $nu{m_{\rm{t}}}$ is the number of traversable patches and ${E_{{\rm{avg}}}}$ is the average ${E}$ of the traversable patches.
		\end{tablenotes} 
	\end{threeparttable}
\end{table}

\begin{table*}[t]
	\centering
	\caption{Methods Comparison}
	\label{table3}
	\begin{tabular}{cc|cccccc}
		\toprule
		\textit{Scenario} & \textit{Method} & \textbf{${T_{\rm{c}}}(\mathop{\rm ms})$} & \textbf{${T_{\rm{p}}}(\mathop{\rm ms})$} & \textbf{${T_{\rm{c}}}+{T_{\rm{p}}}(\mathop{\rm ms})$} & \textbf{$L(\mathop{\rm m})$} & $\kappa ({\mathop{\rm m}^{ - 1}})$ & $P$ \\ \midrule
		&Wang's [2] & 2580.69 & 1370.33 & 3591.02 & 395.01 & 0.109 &  1.0\\
		\textit{Spiral}  &Yang's [3] & 501.98 & 333.25 & 835.23 & \textbf{372.05} & 0.095 &  1.0\\
		&Proposed & \textbf{55.21} & \textbf{201.02} & \textbf{256.23} & 390.89 & \textbf{0.089} &  1.0\\ \midrule
		&Wang's [2] & 11762.51 & 2967.22 & 14729.73 & 363.83 & 0.110 &  0.86\\
		\textit{Uneven terrain} &Yang's [3] & 1303.22 & \textbf{313.58} & 1616.80 & 351.90 & 0.091 &  0.91\\
		&Proposed & \textbf{129.30} & 346.27 & \textbf{475.57} & \textbf{346.72} & \textbf{0.063} &  \textbf{0.95}\\ \midrule
		&Wang's [2] & 2749.84 & 1179.87 & 3929.71 & 209.17 & 0.322 &  0.71\\
		\textit{Building} &Yang's [3] & 798.74 & 339.65 & 1138.39 & \textbf{196.74} & 0.291 &  0.75\\
		&Proposed & \textbf{150.02} & \textbf{249.56} & \textbf{399.58} & 201.62 & \textbf{0.246} &  \textbf{0.91}\\ 
		\bottomrule
	\end{tabular}
\end{table*}
\begin{figure}[t]
	\centering
	\includegraphics[width=\columnwidth]{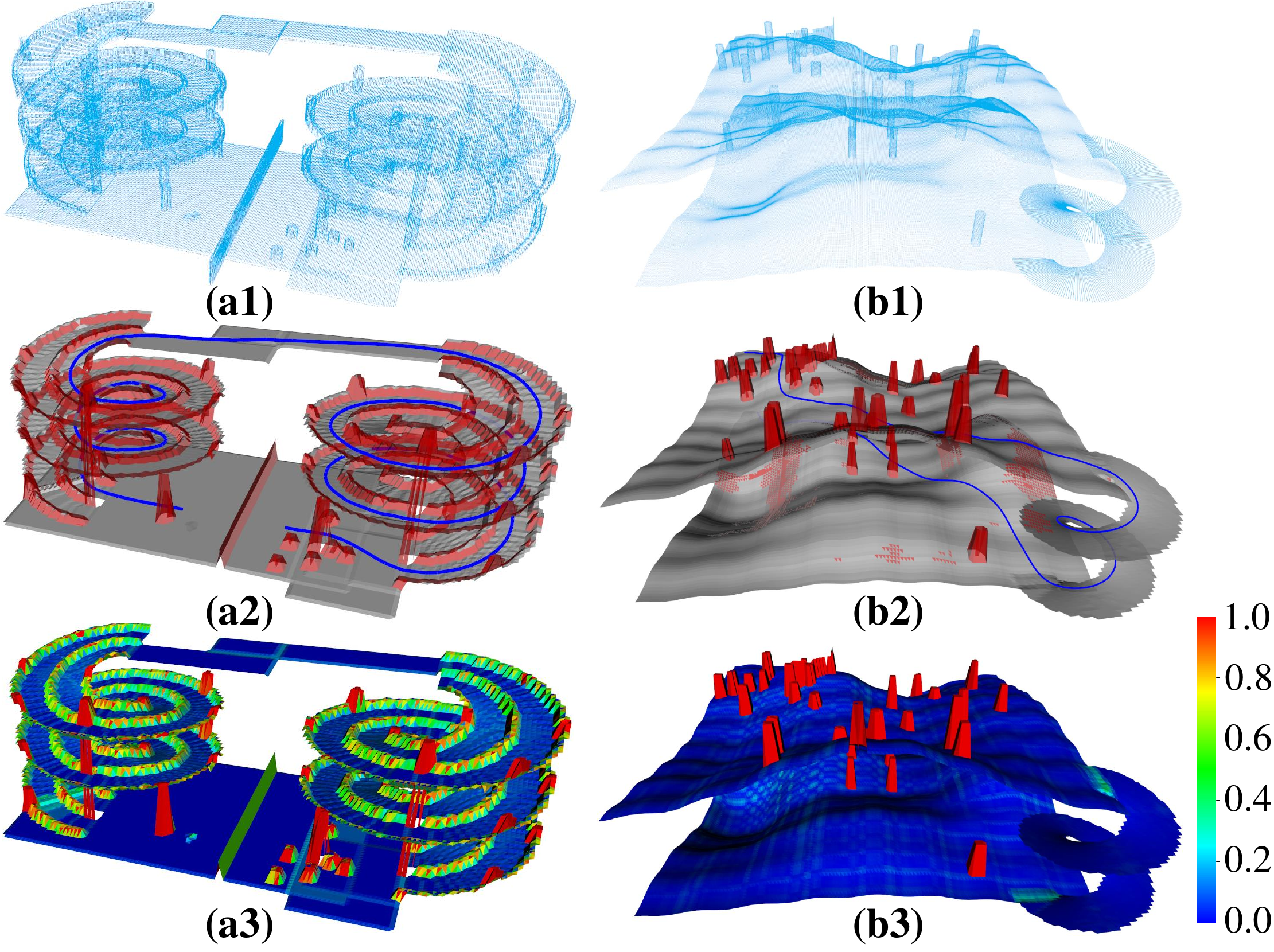}
	\caption{The first row represents the point clouds of the scenarios. The second row represents the results of proposed map construction and trajectory generation methods. The third row represents the ${E}/{E_{{\rm{max}}}}$ of the patches, where ${E_{{\rm{max}}}}$ is the maximum ${E}$ among a scenario.}
	\label{figurelabe7}
\end{figure}

\begin{figure}[t]
	\centering
	\includegraphics[width=\columnwidth]{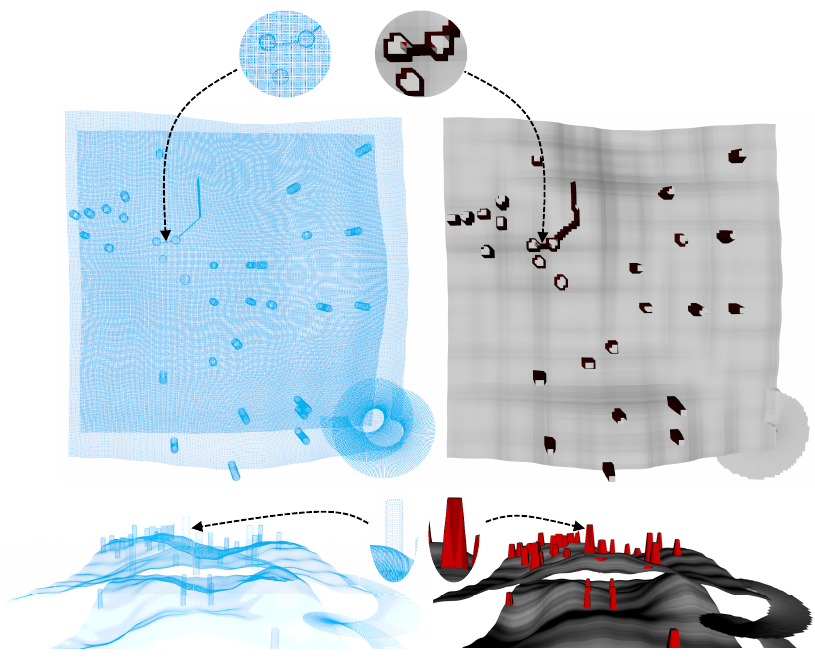}
	\caption{The inflation effect of prposed map construction method.}
	\label{figurelabe7}
\end{figure}

\begin{figure}[h]
	\centering
	\includegraphics[width=\columnwidth]{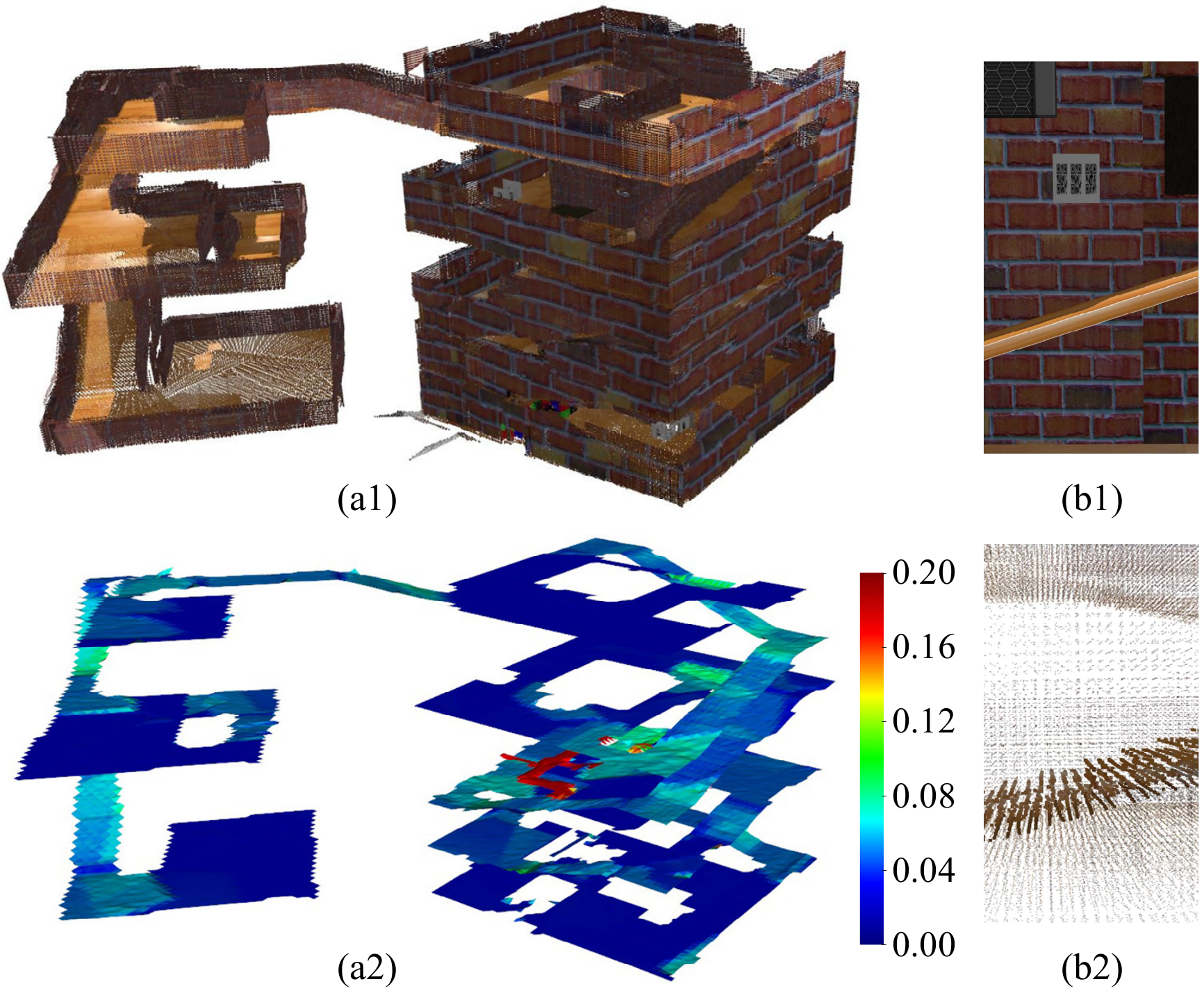}
	\caption{ (a1) represents the point cloud of the building scenario. (a2) represents the ${E}$ of the traversable patches. (b1) represents a slope in the building scenario. (b2) represents the point cloud corresponding to the slope shown in (b1). Due to the inaccuracy in pose estimation, the number of point cloud layers corresponding to the slope exceeds two (the upper and lower surfaces of the slope).}
	\label{figurelabe8}
\end{figure}
\begin{figure}[t]
	\centering
	\includegraphics[width=\columnwidth]{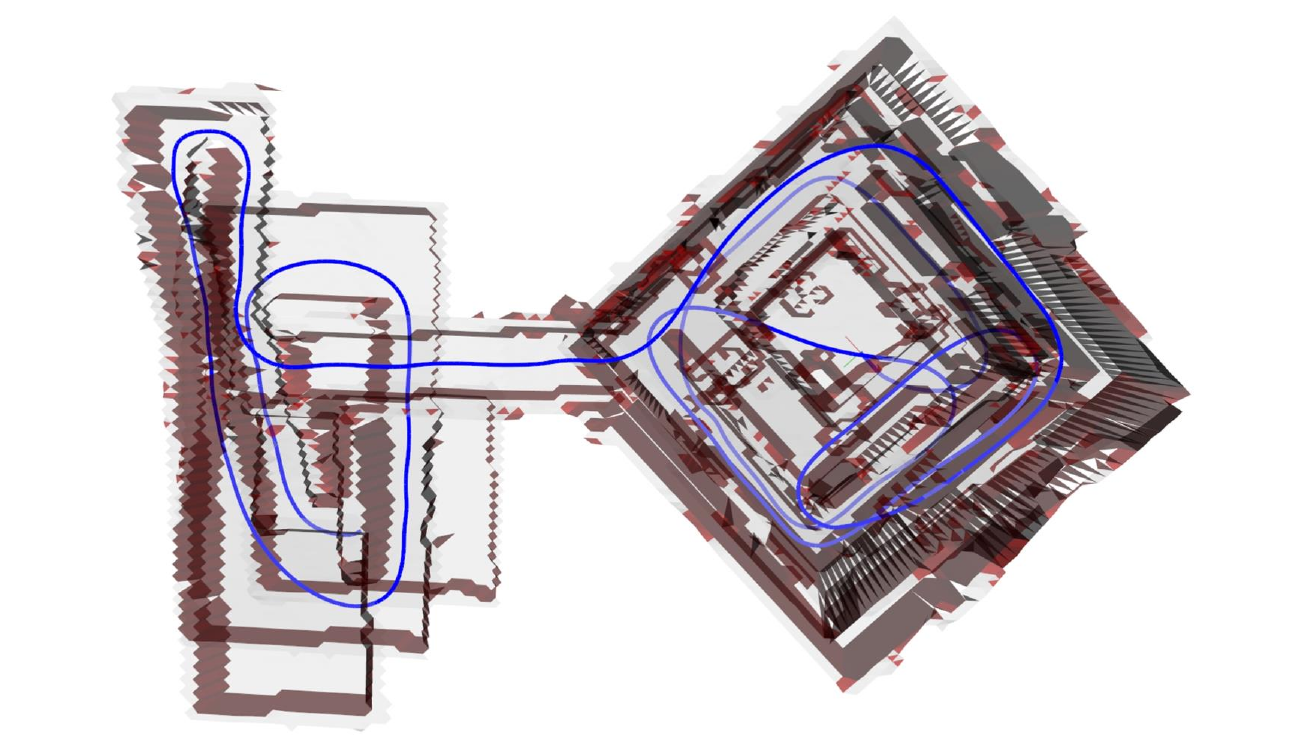}
	\caption{The results of map construction and trajectory generation in the \textit{Building} scenario by proposed method.}
	\label{figurelabe9}
\end{figure}

\begin{figure}[htbp]
	\centering
	\includegraphics[width=\columnwidth]{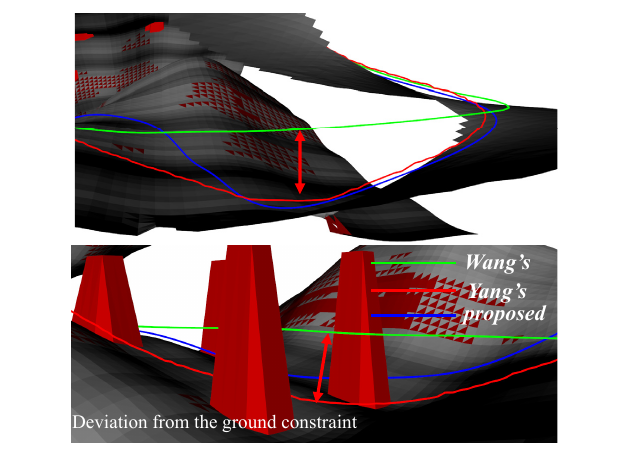}
	\caption{ The deviation from the ground constraint in terrain with high undulation frequency.}
	\label{figurelabe7}
\end{figure}

\subsection{Evaluation of Map Construction}
Given a patch and points set ${\bf{P}}$ located within the corresponding projection area at the same level, the vertical projection discrepancy is calculated by:
$$
{\left. {{E} = \frac{1}{N}\sum\limits_{i = 1}^N {|{f_{\rm{p}}}({x_i}{\rm{,}}{y_i}) - {z_i}|} } \right|_{{{\bf{p}}_i} = ({x_i}{\rm{,}}{y_i}{\rm{,}}{z_i}),{{\bf{p}}_i} \in {\bf{P}}}}\eqno{(14)}
$$
where ${f_{\rm{p}}}$ represents the function describing the plane of patch. For the \textit{Building} scenario, ${\bf{P}}$ is a set of sampled points reflecting the scenario's true heights.

As shown in Fig. 6 (a3, b3), Fig. 8 (a2) and Table II, the traversable patches can represent the scenarios' structures with high accuracy. Since the proposed map construction method selects the highest point as the representative point in regions with steep slope variation and the obstacle inflation effect, the patches' heights being greater than the true heights. However, the proposed trajectory generation method only generates motion primitives on traversable patches which implies the trajectory generated can adapt to terrain condition. 
\subsection{Simulation and Benchmark Comparison}
We deploy the proposed method in these scenarios and compare it with the methods of Wang's \cite{wang2023towards} and Yang's \cite{yang2024efficient}, Fig. 6 (a2, b2), Fig. 9 and Fig. 1 present the results of our method.

All the methods are evaluated on a laptop with an AMD R9-7945hx @ 2.5 GHz.

100 sets of start and goal points are used to evaluate map construction time ${T_{\rm{c}}}$, trajectory generation time ${T_{\rm{p}}}$, trajectory length $L$, and curvature $\kappa$. Subsequently, 100 sets of randomly distributed start and goal points are used to evaluate trajectory generation success rate $P$. The results are summarized in Table III. 

Our method demonstrates a significant advantage in time (${T_{\rm{c}}}$ $+$ ${T_{\rm{p}}}$) efficiency across all three scenarios. In the \textit{Spiral} scenario (Fig. 11 (a)), the trajectory generated by our method can stay in the middle of the spiral roadways. Although the trajectory length is longer than Yang's method \cite{yang2024efficient}, our trajectory is smoother. Wang's method \cite{wang2023towards} produces a suboptimal trajectory towards the end. In the \textit{Uneven terrain} scenario (Fig. 11 (b)), Wang's method \cite{wang2023towards} uses the ESDF map defined in the grid map for trajectory optimization, which pushes 3D trajectory away from objects in any direction. This causes the trajectory to be pushed away from the supporting ground, resulting in deviation from ground constraints, especially in terrain with high undulation frequency (Fig.10). Our method, however, adjusts the waypoints' $z$ values during optimization based on ground constraints, allowing it to adapt well to the terrain condition.

Wang's \cite{wang2023towards} and Yang's\cite{yang2024efficient} methods consider only the topmost point cloud within each level during map construction, making them sensitive to noise. In contrast, our map construction method accounts for all point cloud within each level, resulting in better robustness against noise. Yang's method \cite{yang2024efficient} calculates height gradients using neighboring voxels to evaluate terrain traversability. In areas with significant noise, larger size of voxels and smaller inflation kernel are required to mitigate its impact, which reduces the accuracy of the map representation and consequently leads to a suboptimal trajectory. In the \textit{Building} scenario (Fig. 11 (c)), the trajectory generated by Yang's method \cite{yang2024efficient} is close to obstacles, resulting in a shorter path, while not smooth enough. Wang's method \cite{wang2023towards} causes a wall collision due to the noise. Our method generates a smoother and safer trajectory with higher success rate and time efficiency.
\begin{figure}[htbp]
	\centering
	\includegraphics[width=\columnwidth ]{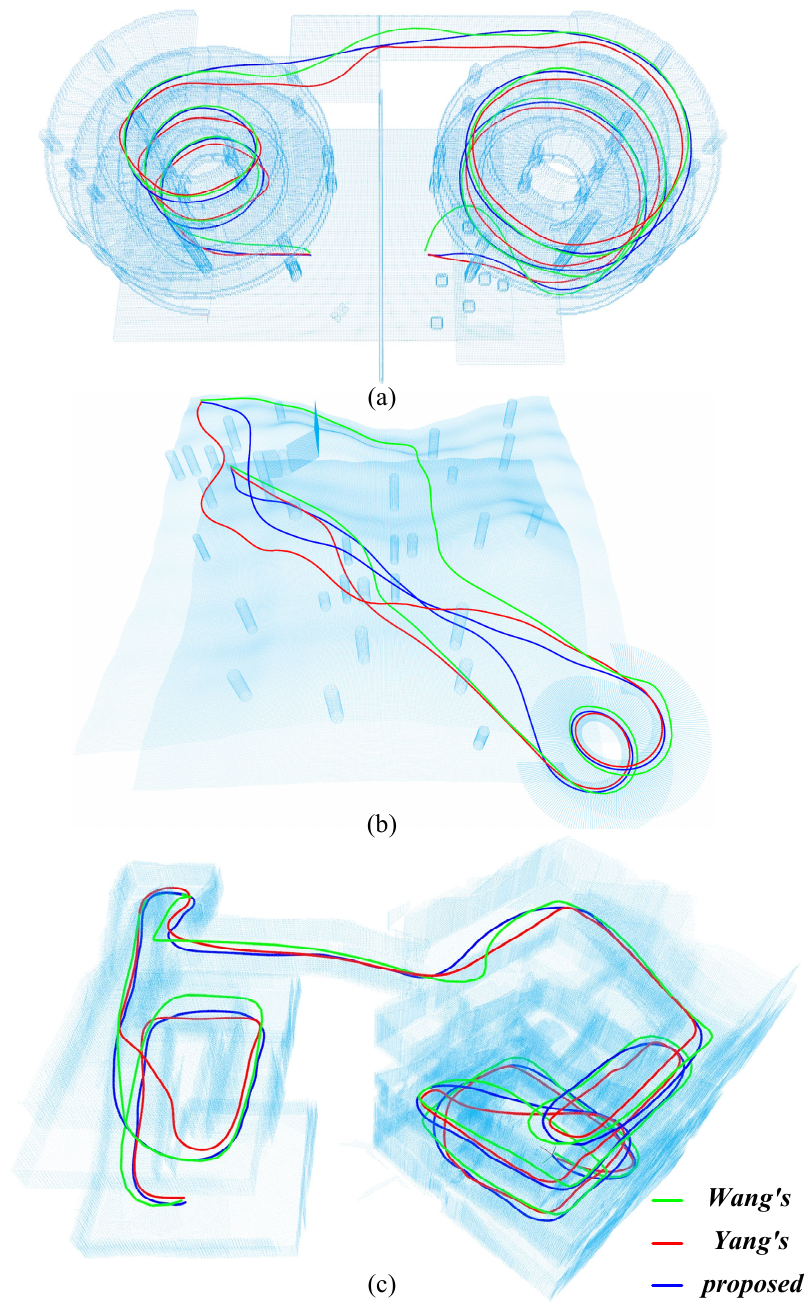}
	\caption{The trajectories of three methods in three scenarios.}
	\label{figurelabe10}
\end{figure}
\section{Conclusions}

In this paper, we proposed a method to construct multi-level map using regularly arranged  patches for 3D environments with complex structures. Based on this map's efficient indexing and continuity, a kinematic path search method is proposed to generate trajectory. The transition between different patches is smooth during primitives generation, ensuring the smoothness of global trajectory. We evaluated the accuracy of the constructed maps and the quality of the generated trajectories in different scenarios, demonstrating that our method can generate smooth and safe trajectories while maintaining high computational efficiency and robustness.

In the future, our research will combine the dynamic model of the robot's navigation in 3D environment with the trajectory generation method. Additionally, exploration-based navigation without prior knowledge of global point cloud will be taken into consideration.





\bibliographystyle{IEEEtran}
\bibliography{IEEEabrv,References}
\end{document}